\documentclass{svproc}
\usepackage[utf8]{inputenc}

\usepackage{url}

\usepackage{amssymb}

\usepackage{multicol}

\usepackage[figuresright]{rotating}

\usepackage[table]{xcolor}

\definecolor{Gray}{gray}{0.85}

\usepackage{colortbl}

\usepackage[inkscapearea=page]{svg}
\usepackage{adjustbox}

\usepackage{graphicx}

\usepackage[misc,geometry]{ifsym}

\usepackage{tikz}
\usepackage{scalerel}
\usetikzlibrary{svg.path}
\definecolor{orcidlogocol}{HTML}{A6CE39}
\tikzset{
  orcidlogo/.pic={
    \fill[orcidlogocol] svg{M256,128c0,70.7-57.3,128-128,128C57.3,256,0,198.7,0,128C0,57.3,57.3,0,128,0C198.7,0,256,57.3,256,128z};
    \fill[white] svg{M86.3,186.2H70.9V79.1h15.4v48.4V186.2z}
                 svg{M108.9,79.1h41.6c39.6,0,57,28.3,57,53.6c0,27.5-21.5,53.6-56.8,53.6h-41.8V79.1z M124.3,172.4h24.5c34.9,0,42.9-26.5,42.9-39.7c0-21.5-13.7-39.7-43.7-39.7h-23.7V172.4z}
                 svg{M88.7,56.8c0,5.5-4.5,10.1-10.1,10.1c-5.6,0-10.1-4.6-10.1-10.1c0-5.6,4.5-10.1,10.1-10.1C84.2,46.7,88.7,51.3,88.7,56.8z};
  }
}

\newcommand\orcidicon[1]{\href{https://orcid.org/#1}{\mbox{\scalerel*{
\begin{tikzpicture}[yscale=-1,transform shape]
\pic{orcidlogo};
\end{tikzpicture}
}{|}}}}

\makeatletter
\let\NAT@parse\undefined
\makeatother
\usepackage{hyperref}
\hypersetup{
    bookmarks=false,
    colorlinks=true,
    citecolor=blue,
    linkcolor=blue,
    filecolor=blue,
    hypertexnames=true,
    urlcolor=blue,
    pagebackref=true,
    pdftitle={Springer},
    pdfpagemode=FullScreen
    }
\begin{document}
\mainmatter              
\title{Real-Time And Robust 3D Object Detection with Roadside LiDARs}

\titlerunning{Real-Time And Robust 3D Object Detection with Roadside LiDARs}  
%

\author{Walter Zimmer$^{\textrm{(\Letter)}}$~\orcidicon{0000-0003-4565-1272} \and Jialong Wu~\orcidicon{0000-0003-3875-648X} \and Xingcheng Zhou~\orcidicon{0000-0003-1178-5221}\and\\Alois~C. Knoll~\orcidicon{0000-0003-4840-076X}}
\authorrunning{Walter Zimmer et al.} 
%
\tocauthor{Walter Zimmer, Jialong Wu, Xingcheng Zhou, Alois C. Knoll}
\institute{School of Computation, Information and Technology, Department of Computer Engineering, Technical University of Munich (TUM), 85748 Garching, Germany,\\
 \email{\{walter.zimmer,jialong.wu,xingcheng.zhou\}@tum.de, knoll@in.tum.de},\\
 \texttt{https://www.ce.cit.tum.de/air}
}

\maketitle              

\begin{abstract}
This work aims to address the challenges in autonomous driving by focusing on the 3D perception of the environment using roadside LiDARs. We design a 3D object detection model that can detect traffic participants in roadside LiDARs in real-time. Our model uses an existing 3D detector as a baseline and improves its accuracy. To prove the effectiveness of our proposed modules, we train and evaluate the model on three different vehicle and infrastructure datasets. To show the domain adaptation ability of our detector, we train it on an infrastructure dataset from China and perform transfer learning on a different dataset recorded in Germany. We do several sets of experiments and ablation studies for each module in the detector that show that our model outperforms the baseline by a significant margin, while the inference speed is at 45\,Hz (22\,ms).  We make a significant contribution with our LiDAR-based 3D detector that can be used for smart city applications to provide connected and automated vehicles with a far-reaching view. Vehicles that are connected to the roadside sensors can get information about other vehicles around the corner to improve their path and maneuver planning and to increase road traffic safety.

\end{abstract}
\section{Introduction}
\subsection{Problem Statement}
High quality and balanced data is crucial to achieve high accuracy in deep learning applications. We analyze and split the task into four challenges. The first challenge is the lack of labeled frames of roadside LiDARs. Considering the high labor cost of manual labeling for 3D bounding boxes in LiDAR point clouds, we need to find a solution to deal with small datasets and use only few labeled frames for training. Publicly available LiDAR datasets were recorded and labeled from a driver perspective which makes is difficult to apply these datasets for roadside LiDARs. The second research question this work is dealing with lies in the area of domain adaptation. How can a neural network that was trained in one operational design domain (ODD), e.g. on data recorded by vehicle sensors like in the \textit{KITTI} dataset \cite{geiger2012we,geiger2013vision}, be adapted to a different domain (e.g. to data recorded by roadside sensors on a highway or a rural area? Much research has been done in the area of domain adaptation and transfer learning \cite{triess2021survey,zhang2021srdan} --- training a model on a large dataset (source domain) and fine-tuning it on a smaller dataset (target domain). Another challenge is real-time 3D object detection on roadside LiDARs, i.e. to detect objects with a high enough frame rate to prevent accidents. This highly depends on the LiDAR type and the number of points per scan that need to be processed. The final challenge this work is dealing with is a robust 3D detection of all traffic participants. Detecting small and occluded objects in different weather conditions and rare traffic scenarios is a highly important research area to increase safety of automated vehicles.\\

\begin{figure}[t]
    \centering
    \adjustbox{trim=0 1cm 0 0}{
        \includegraphics[width=\linewidth]{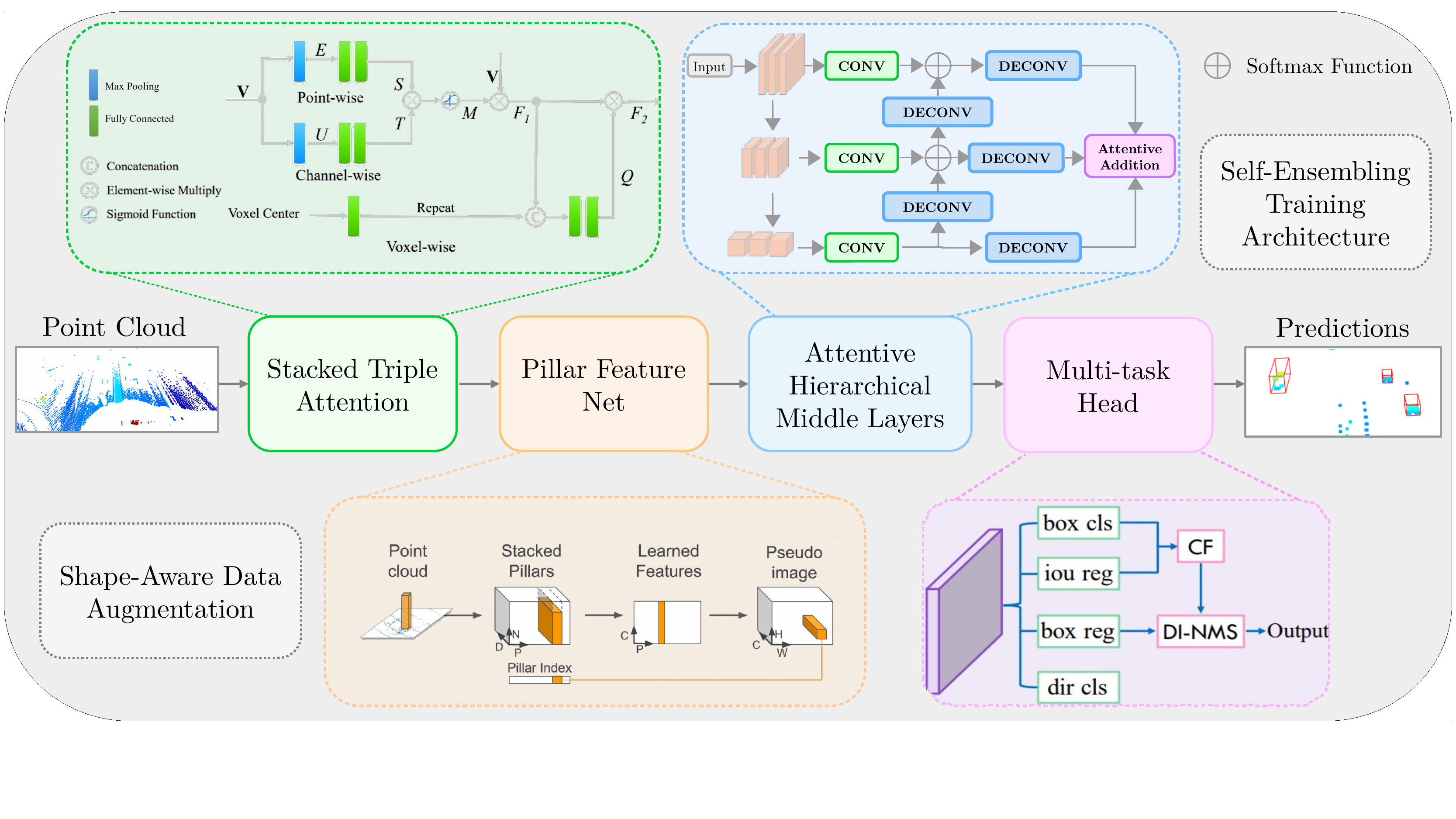}
    }
    \caption{Overview architecture of SE-ProPillars, a LiDAR-only single-stage pillar-based 3D object detector. The detector is based on PointPillars \cite{lang2019pointpillars}, with the following five additional extensions. 1) Shape-Aware Data Augmentation \cite{zheng2021se}, a training technique to improve the accuracy without adding any additional costs in the inference time. 2) The Stacked Triple Attention Mechanism \cite{liu2020tanet} enhances the learned features from the raw point cloud using the triple attention mechanism, including channel-wise, point-wise, and voxel-vise attention. 3) The pillar feature net turns point-wise features into pillar features and scatters the pillar features into a pseudo image. 4) We propose the Attentive Hierarchical Middle Layers to perform 2D convolution operations on the pseudo image. Hierarchical feature maps are concatenated with an attentive addition operation. 5) The Multi-task detection head \cite{zheng2020cia} is used for the final prediction, that includes an IoU prediction to alleviate the misalignment between the localization accuracy and classification confidence. 6) Finally, a Self-Ensembling Training Architecture (a teacher \& student training framework) \cite{zheng2021se} is used as a training technique in the training process that leads to a substantial increase in the precision without affecting the inference time.}
    \label{fig:architecture}
\end{figure}

\subsection{Objectives}
\begin{figure}[t]
    \centering
    \includegraphics[height=4cm]{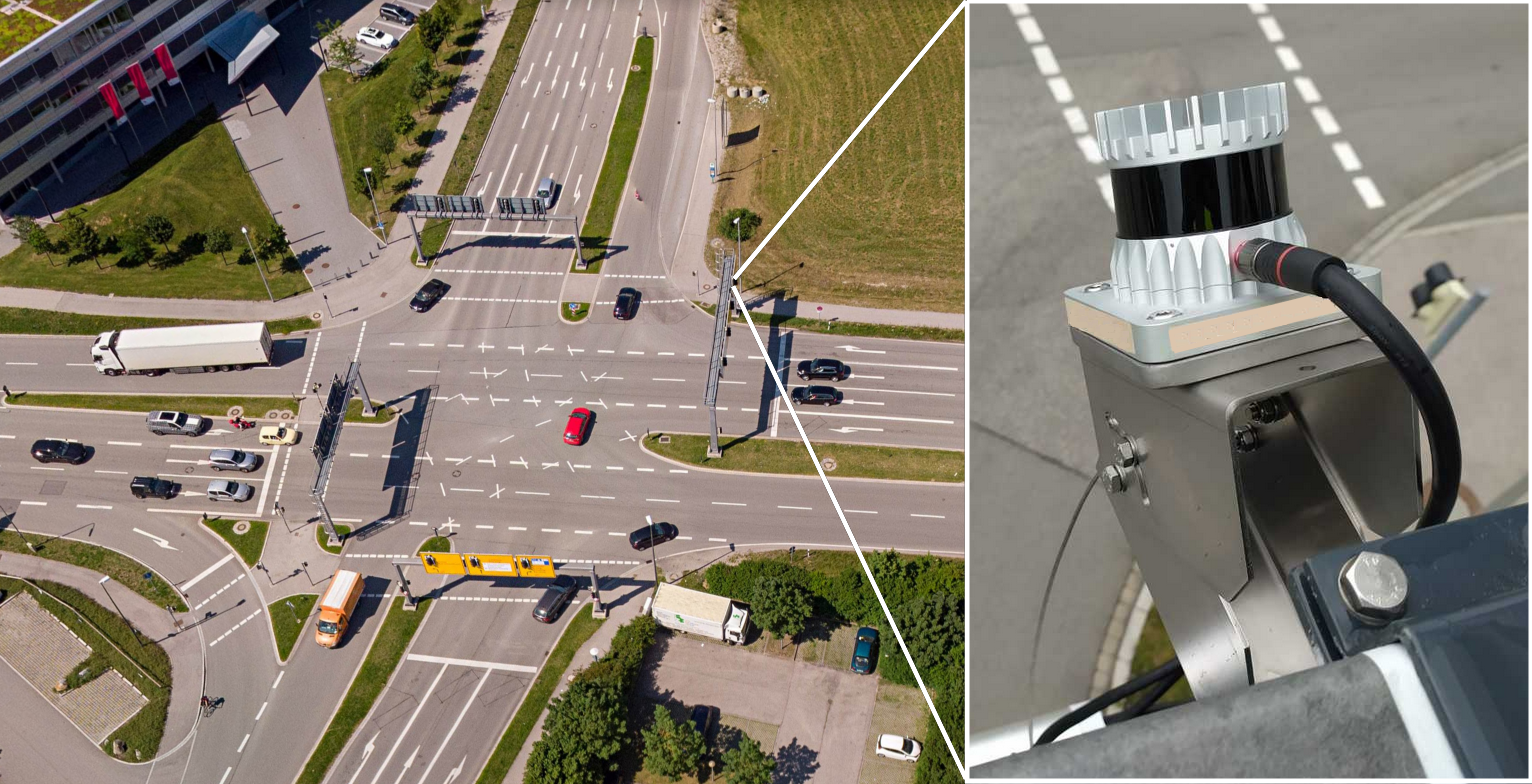}
    \includegraphics[height=4cm]{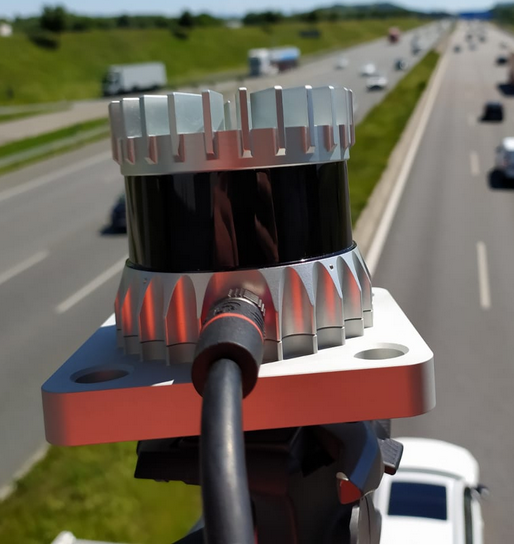}
    \caption{Left: Intersection where 3D Object Detection with roadside LiDARs is performed. Right: Roadside LiDAR mounted above the Autobahn A9.}
    \label{fig:intersection}
\end{figure}

The first objective is to create a large infrastructure dataset with sufficient labeled point cloud frames. This dataset should be balanced in terms of object classes and contain a high variety so that objects can be detected in different scenarios and different environment conditions. Another objective is to analyze whether transfer learning from a larger roadside LiDAR dataset, such as the recently released \textit{IPS300+} dataset \cite{wang2021ips300+}, can improve the model performance. The first batch of the published IPS300+ dataset includes 1,246 labeled point cloud frames and contains on average 319.84 labels per frame. In this work the aim is to design a single-stage 3D object detector that can detect objects in real-time using roadside LiDARs. An example of an intersection that is equipped with LiDARs is shown in Fig. \ref{fig:intersection}. The goal is to reach an inference rate of at least 25\,Hz, which leads to a maximum inference time of 40\,ms per point cloud frame. In terms of accuracy the target is to achieve at least 90\% mean average precision (mAP) on the Car class within the test set. In the end the designed model needs to reach a reasonable trade-off between the inference time and the model performance. The final target is to evaluate the model performance and speed on the first released batch of the manually labeled \textit{A9-Dataset} \cite{cress2022a9} and on publicly available infrastructure datasets. The \textit{A9-Dataset} was recorded on the Providentia++ Test Stretch in Munich, Germany \cite{krammer2019providentia}.\\

\subsection{Contribution}
The work provides a complete analysis and solution for the single LiDAR detection task. The main contributions in this work can be summarized as follows:
\begin{itemize}
    \item We propose a single-stage LiDAR-only detector based on PointPillars. We introduce five extensions to improve PointPillars and evaluate the performance of the designed model on KITTI's validation set and the test set of IPS300+ and A9-Dataset to check its validity.
    \item We design a point cloud registration algorithm for LiDARs installed on the infrastructure to increase the point density.
    \item We create a synthetic dataset, called proSynth, with 2k labeled LiDAR point cloud frames using the CARLA simulator \cite{dosovitskiy2017carla} and train our model on that.
    \item Experiments show that our \textit{SE-ProPillars} model outperforms the \textit{PointPillars} model by (+4.20\%,+1.98\%,+0.10\%) 3D mAP on easy, moderate and hard difficulties respectively, while the inference speed reaches 45\,Hz (22\,ms).
    \item We train on the IPS300+ dataset, perform transfer learning on A9-Dataset, and achieve a $3D~mAP@0.5$ of 50.09\% on the Car class of the A9 test set using 40 recall positions and a NMS IoU of 0.1.
\end{itemize}

\section{Related Work}
\label{sec:related_work}
 According to the form of feature representation, LiDAR-only 3D object detectors can be divided into four different types, i.e. point-based, voxel-based, hybrid and projection-based methods.

\subsection{Point-based Methods}
In point-based methods, features maintain the form of point-wise features, either by a sampled subset or derived virtual points. PointRCNN \cite{pointrcnn} uses a PointNet++ backbone \cite{pointnet++} to extract point-wise features from the raw point cloud, and performs foreground segmentation. Then for each foreground point, it generates a 3D proposal followed by a point cloud ROI pooling and a canonical transformation-based bounding box refinement process. Point-based methods usually have to deal with a huge amount of point-wise features, which leads to a relatively lower inference speed. To accelerate point-based methods, 3DSSD \cite{3dssd} introduces a feature farthest-point-sampling (F-FPS), which computes the feature distance for sampling, instead of Euclidean distance in traditional distance farthest-point-sampling (D-FPS). The inference speed of 3DSSD is competitive with voxel-based methods.

\subsection{Voxel-based Methods}
VoxelNet \cite{voxelnet} divides the 3D space into equally spaced voxels and encodes point-wise features into voxel-wise features. Then 3D convolutional middle layers operate on these encoded voxel features. 3D convolution on sparse point cloud space brings too much unnecessary computational cost. SECOND \cite{yan2018second} proposes to use a sparse convolutional middle extractor \cite{subsparseCNN,sparseCNN}, which greatly speeds up the inference time. In PointPillars \cite{lang2019pointpillars}, the point cloud is divided into pillars (vertical columns), which are special voxels without any partition along the z-direction. The feature map of pillars can be treated as a pseudo-image, and therefore the expensive 3D convolution is replaced by 2D convolution. PointPillars achieves the fastest speed with the help of TensorRT acceleration. SA-SSD \cite{sassd} adds a detachable auxiliary network to the sparse convolutional middle layers to predict a point-wise foreground segmentation and a center estimation task, which can provide additional point-level supervision. SA-SSD also proposes a part-sensitive warping (PS-Warp) operation as an extra detection head. It can alleviate the misalignment between the predicted bounding boxes and classification confidence maps, since they are generated by two different convolutional layers in the detection head. CIA-SSD \cite{zheng2020cia} also notices the misalignment issue. It designs an IoU-aware confidence rectification module, using an additional convolutional layer in the detection head to make IoU predictions. The predicted IoU value is used to rectify the classification score. By introducing only one additional convolutional layer, CIA-SSD is more lightweight than SA-SSD. SE-SSD \cite{zheng2021se} proposes a self-ensembling post-training framework, where a pre-trained teacher model produces predictions that serve as soft targets in addition to the hard targets from the label. These predictions are matched with student's predictions by their IoU and supervised by the consistency loss. Soft targets are closer to the predictions from the student model and therefore can help the student model to fine-tune its predictions. The Orientation-Aware Distance-IoU Loss is proposed to replace the traditional smooth-$L_1$ loss of bounding box regression in the post training, in order to provide a fresh supervisory signal. SE-SSD also designs a shape-aware data augmentation module to improve the generalization ability of the student model.

\subsection{Hybrid Methods}
Hybrid methods aim to take advantage of both point-based and voxel-based methods. Point-based methods have a higher spatial resolution but involve higher computational cost, while voxel-based methods can efficiently use CNN layers for feature extraction but lose local point-wise information. Hybrid methods try to strike a balance between them.

HVPR \cite{hvpr} is a single-stage detector. It has two feature encoder streams extracting point-wise and voxel-wise features. Extracted features are integrated together and scattered into a pseudo image as hybrid features. An attentive convolutional middle module is performed on the hybrid feature map, followed by a single-stage detection head. STD \cite{std} is a two-stage detector that uses PointNet to extract point-wise features. A point-based proposal generation module with spherical anchors is designed to achieve high recall. Then a PointsPool module voxelizes each proposal, followed by a VFE layer. In the box refinement module, CNNs are applied on those voxels for final prediction. PV-RCNN \cite{pvrcnn} uses the 3D sparse convolution for voxel feature extraction. A Voxel Set Abstraction (Voxel-SA) module is added to each convolutional layer to encode voxel features into a small set of key points, which are sampled by farthest point sampling. Key point features are then re-weighted by foreground segmentation score. Finally, they are used to enhance the ROI grid points for refinement. H$^2$3D R-CNN \cite{h23drcnn} extracts point-wise features from multi-view projection. It projects the point cloud to a bird's eye view under Cartesian coordinates and a perspective view under the cylindrical coordinates, separately. BEV feature and point-voxel (PV) features are concatenated together for proposal generation in BEV and fused together as point-wise hollow-3D (H3D) features. Then voxelization is performed on the 3D space, and point-wise H3D features are aggregated as voxel-wise H3D features for the refinement process.

\subsection{Projection-based Methods}
RangeDet \cite{fan2021rangedet} is an anchor-free single-stage LiDAR-based 3D object detector that is purely based on the range view representation. It is compact and has no quantization errors. The inference speed is 12\,Hz using a RTX 2080 TI GPU. The runtime is not affected by the expansion of the detection distance, whereas a BEV representation will slow down the inference time as the detection range increases. RangeRCNN \cite{liang2020rangercnn} is another 3D object detector that uses a range image, point view and bird's eye view (BEV). The anchor is defined in the BEV to avoid scale variation and occlusion. In addition, a two-stage regional convolutional neural network (RCNN) is used to improve the 3D detection performance.

\section{Data Generation}

\begin{figure}[t]
    \centering
    \includegraphics[height=4cm]{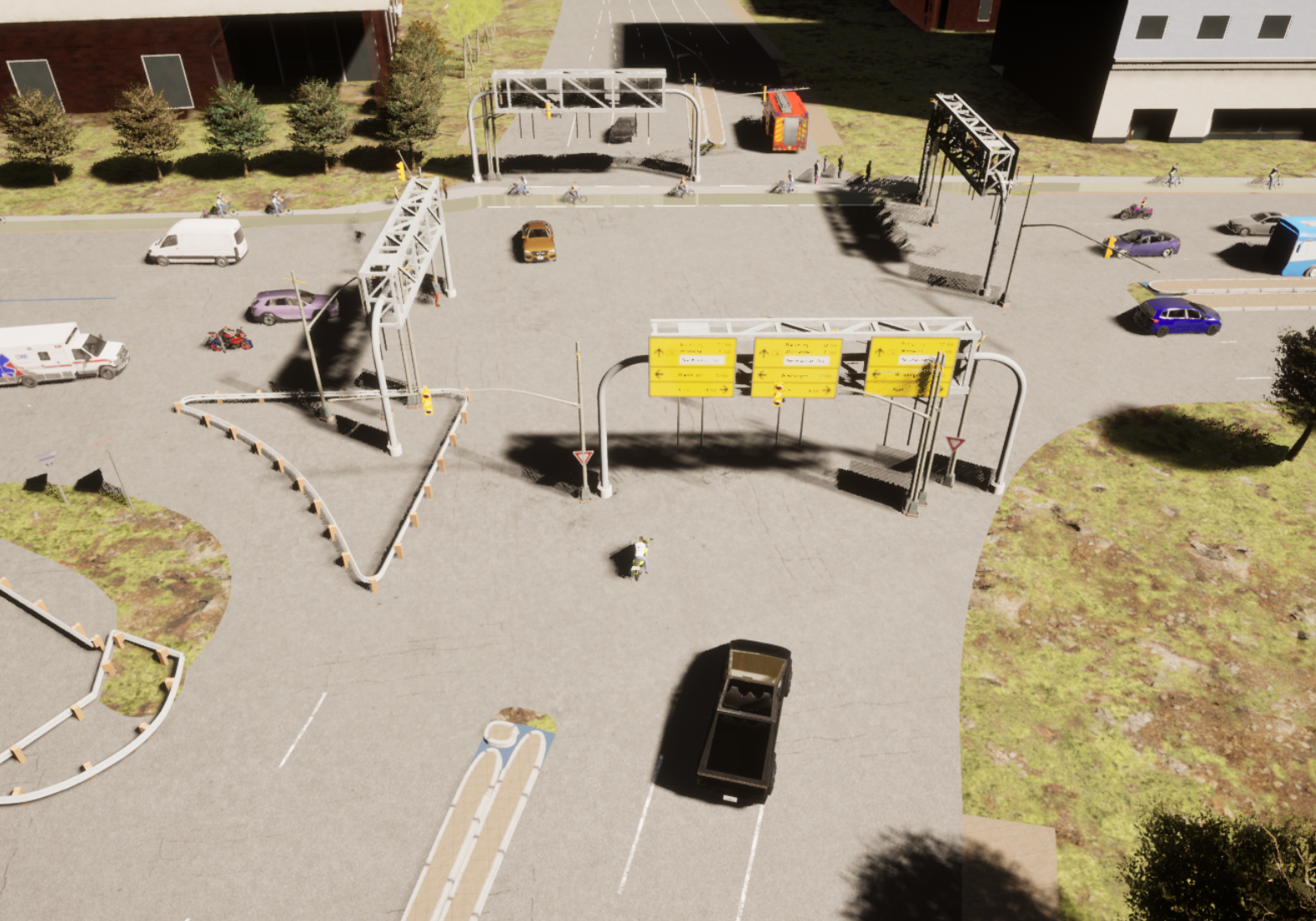}
    \includegraphics[height=4cm]{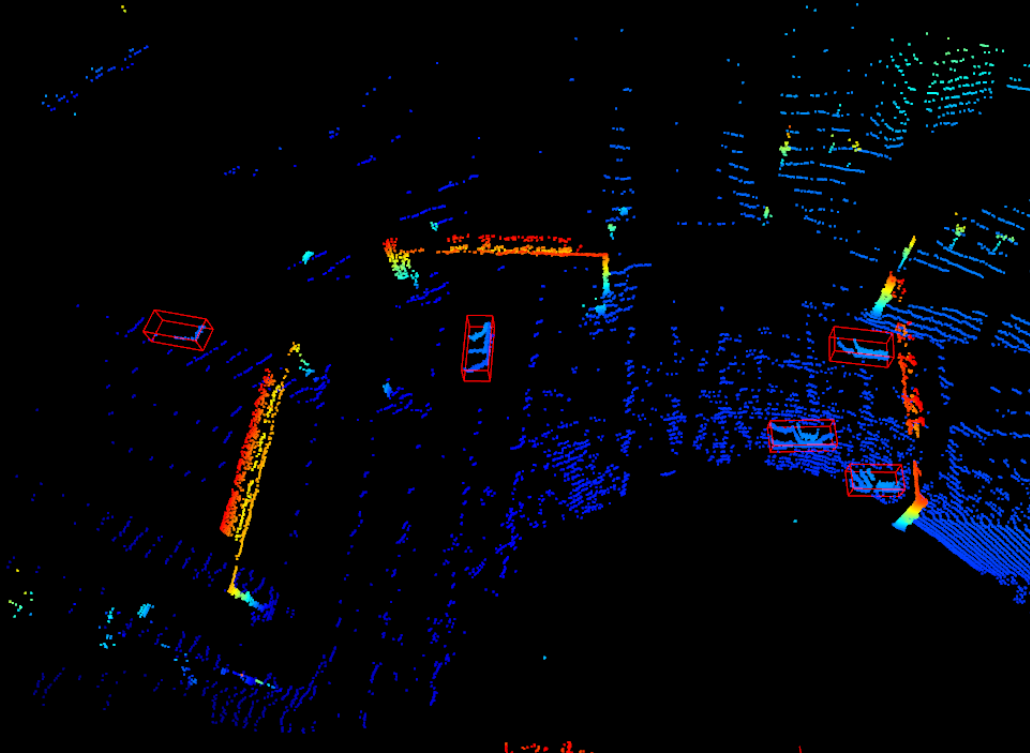}
    \caption{Left: Example of the intersection that is part of the A9 Test Stretch that was modelled in the CARLA simulator. Right: Synthetic point cloud frames with labeled ground truth vehicles extracted from CARLA simulator.}
    \label{fig:intersection_carla}
\end{figure}

\subsection{Real Data Generation}
The first step is to record point cloud data from roadside LiDARs. These recordings should cover a large variety of different scenarios (e.g. traffic jam, overtaking, lane change, lane merge, tail-gate events, accidents, etc.). In the next step (data selection), scenarios of high importance are selected for labeling. The point clouds are then converted into the right format (.pcd) and the ground is removed using the \textit{RANSAC} algorithm. Dynamic objects are then labeled using a custom 3D point cloud labeling tool called \textit{proAnno}, that is based on the open source 3D bounding box annotation tool \textit{3D-BAT} \cite{zimmer20193d}. 456 LiDAR frames were manually labeled to form the first batch of the A9-Dataset.

\subsection{Synthetic Data Generation}
 We created a synthetic dataset (proSynth) with 2000 point cloud frames using the CARLA simulator and train our \textit{SE-ProPillars} model on it. Figure \ref{fig:intersection_carla} shows an example of an intersection with generated traffic in the CARLA simulator. A simulated LiDAR sensor represents a real Ouster OS1-64 (gen. 2) LiDAR sensor with 64 channels and a range of 120 m. In the simulation a noise model is used with a standard deviation of 0.1 to disturb each point along the vector of its raycast. The LiDAR emits 1.31 million points per second and runs at 10\,Hz (131,072 points per frame). The extracted labels were stored in .json files according to the OpenLABEL standard \cite{croce2021open}.

\section{Approach}
\label{sec:approach}
We design a real-time LiDAR-only 3D object detector (\textit{SE-ProPillars}) that could be applied to real-world scenarios. The architecture of the designed \textit{SE-ProPillars} model is shown in Fig. \ref{fig:architecture}. We describe the each part of the detector in the following sections.

\subsection{Point Cloud Registration}
\begin{figure}[t]
    \centering
    \includegraphics[width=0.49\linewidth]{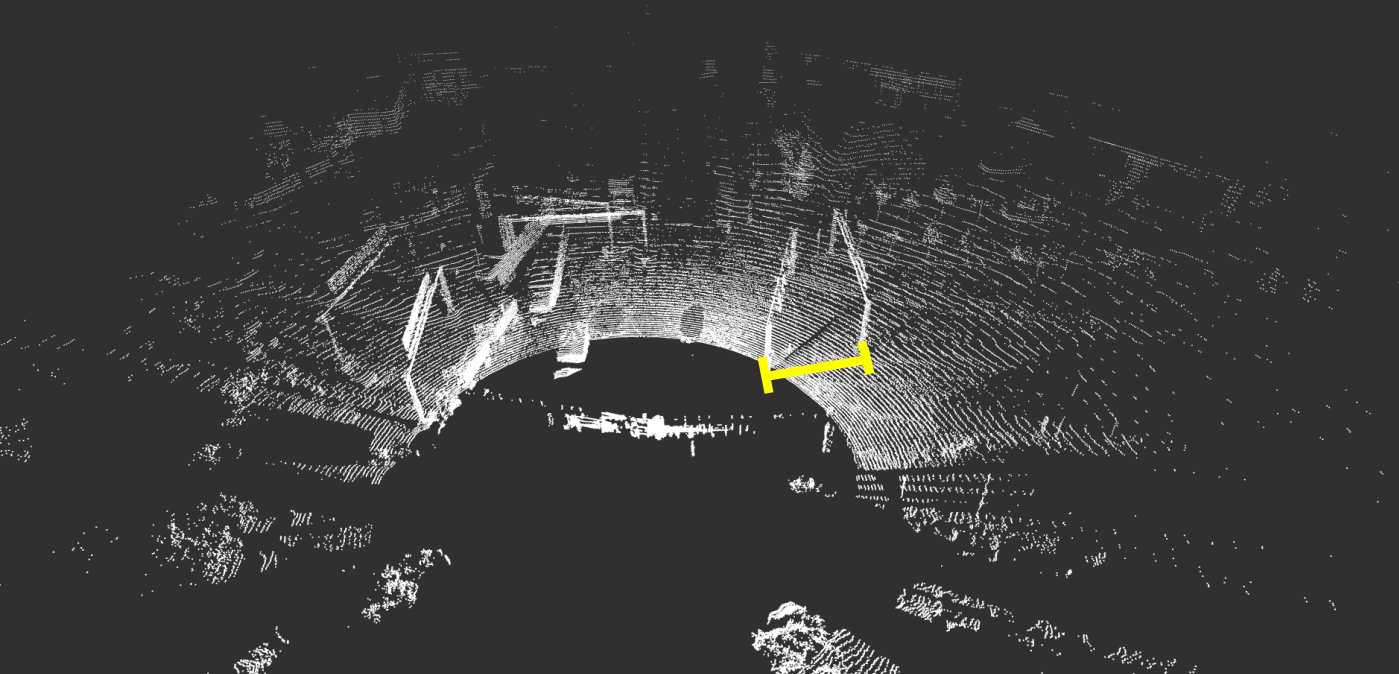}
    \includegraphics[width=0.49\linewidth]{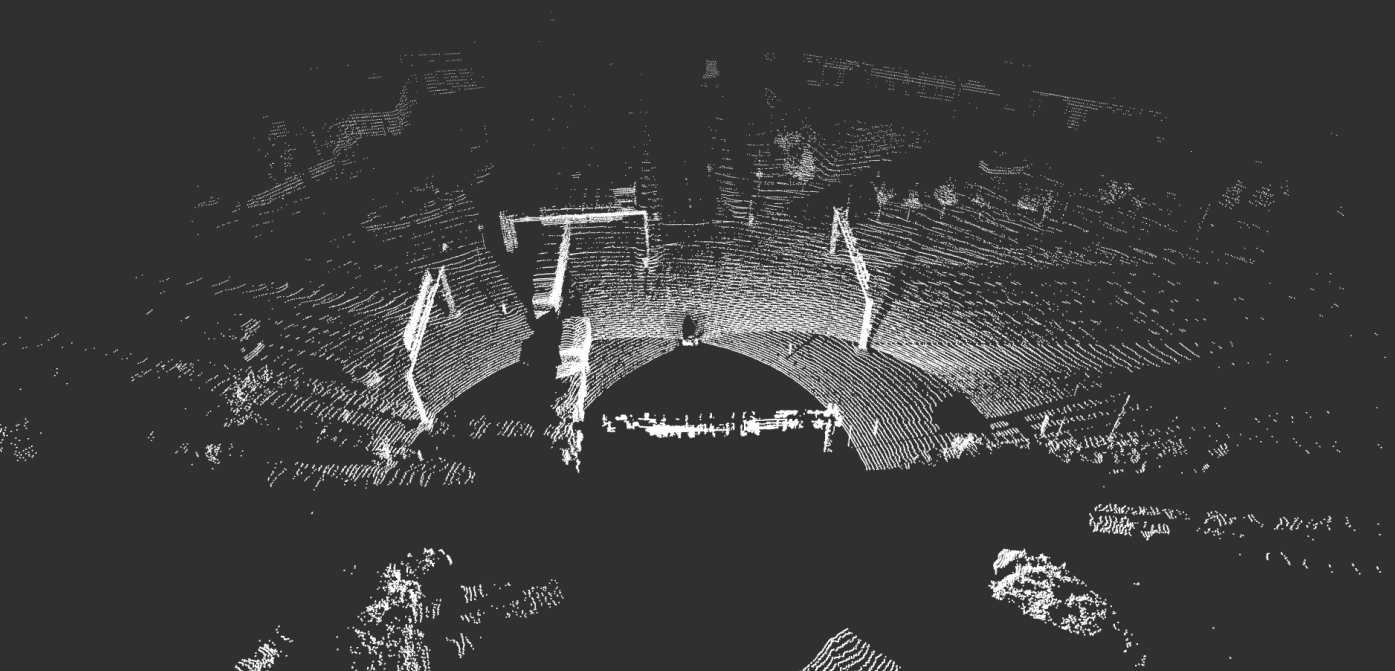}
    \caption{Left: Two point cloud scans before registration. The displacement error is marked in yellow. Right: Two point cloud scans after registration.}
    \label{fig:registration}
\end{figure}

We first design a point cloud registration algorithm for LiDARs installed on the infrastructure to increase the point density and facilitate the detection task. The two Ouster OS1-64 (gen. 2) LiDARs are mounted side by side (about 13\,m apart from each other) on the gantry bridge. Both LiDARs are time synchronized using the ROS time that itself is synchronized with an NTP time server. One LiDAR is treated as the source ($L_s$) and the other one as target ($L_t$). Both LiDAR sensors capture $N$ point cloud scans and each scan is marked with a Unix timestamp: $A=\{a_i\}_{i=1,...,N}$ and $B=\{b_i\}_{i=1,...,N}$. The goal is to put both point clouds into the same coordinate system, by transforming the source point cloud into the target point cloud's coordinate system. To achieve this a set of correspondence pairs $S=\{s_i\}_{i=1,...,n}$ and $T=\{t_i\}_{i=1,...,n}$ need to be found, where $s_i$ corresponds to $t_i$. Then the rigid transformation \textbf{$T$} that includes the rotation \textbf{$R$} and translation \textbf{$t$} is estimated by minimizing the root-mean-squared error (RMSE) between correspondences. Inspired by \cite{kloeker2020real}, we provide an initial transformation matrix to help the registration algorithms better overcome local optima. The initial transformation was acquired with a Real-time Kinematic (RTK) GPS device. With this initial transformation the continuous registration is less likely to be trapped in a local optimum. The continuous registration is done by point-to-point ICP.
The registration process of two Ouster LiDARs, running at 10\,Hz, takes 18.36\,ms on an Intel Core i7-9750H CPU. A RMSE of 0.52164 could be achieved using a voxel size of 2 m. Figure \ref{fig:registration} shows the point cloud scans before and after registration.

\subsection{Voxelization}
We divide the raw point cloud into vertical pillars before feeding them into a neural network. These are special voxels that are not split along the vertical axis. Pillars have several advantages over voxels. A pillar-based backbone is faster than a voxel-based backbone due to fewer grid cells. Time consuming 3D convolutional middle layers are also being eliminated and instead 2D convolutions are being used. We also do not need to manually tune the bin size along the z-direction hyperparameter. If a pillar contains more points than specified in the threshold, then the points are being subsampled to the threshold using farthest point sampling \cite{eldar1997farthest}. If a pillar contains fewer points than the threshold, then it is padded with zeros to make the dimensions consistent. Due to the sparsity issue most of the pillars are empty. We record the coordinates of non-empty pillars according to the pillar's center. Empty pillars are not being considered during the feature extraction until all pillars are being scattered back to a pseudo image for 2D convolution.

\subsection{Stacked Triple Attention}
The \textit{Stacked Triple Attention} module is used for a more robust and discriminative feature representation. Originally introduced in \textit{TANet} \cite{liu2020tanet} by Liu et. al., this module enhances the learning of objects that are hard to detect and better deals with noisy point clouds. The TA module extracts features inside each pillar, using point-wise, channel-wise, and voxel-wise attention. The attention mechanism in this module follows the Squeeze-and-Excitation pattern \cite{hu2018squeeze}.

\begin{figure}[htbp]
    \centering
    \includegraphics[width=1.0\linewidth]{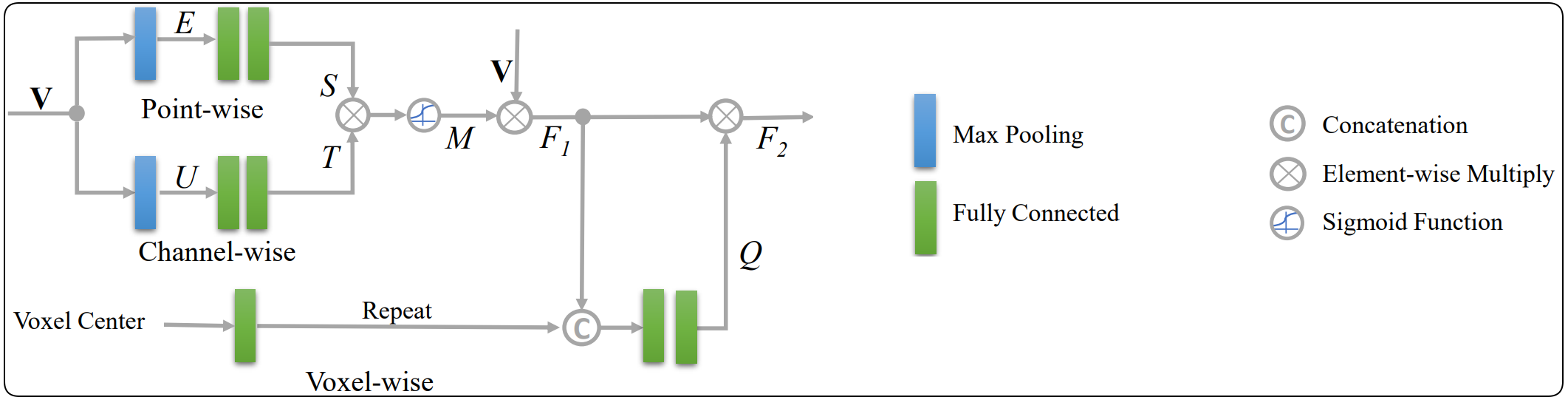}
    \caption{Structure of the triple attention module \cite{liu2020tanet}.}
    \label{fig:triple_attention}
\end{figure}

The structure of the triple attention module is shown in Fig. \ref{fig:triple_attention}. The input $V$ to the module is a $(P \times N \times C)$ tensor, where $P$ is the number of non-empty pillars, $N$ is the maximum number of points in each pillar, and $C$ is the dimension of the input point-wise feature. In the upper branch point-wise attention, following the Squeeze-and-Excitation pattern, we firstly perform max pooling to aggregate point-wise features across the channel-wise dimensions, and then we compute the point-wise attention score $S$ using two fully connected layers. Similarly, the middle branch channel-wise attention aggregates channel-wise features across their point-wise dimensions, to get the channel-wise attention score $T$. Then $S$ and $T$ are combined by element-wise multiplication, followed with a sigmoid function to get the attention scale matrix $M$, $M = \sigma(S \times T)$. $M$ is then multiplied with input $V$, to get the feature tensor $F1$. In the bottom branch voxel-wise attention, the $C$-dim channel-wise feature in $F1$ is enlarged by the voxel center (arithmetic mean of all points inside the pillar) to $C+3$-dim for better voxel-awareness. Then the enlarged $F1$ is fed into two fully connected layers. The two FC layers respectively compress the point-wise and the channel-wise dimensions to 1, to get the voxel-wise attention score. Finally, a sigmoid function generates the voxel-wise attention scale $Q$, multiplied with the original $F1$ to generate the final output of the TA module $F2$.

\begin{figure}[htbp]
    \centering
    \includegraphics[width=0.5\linewidth]{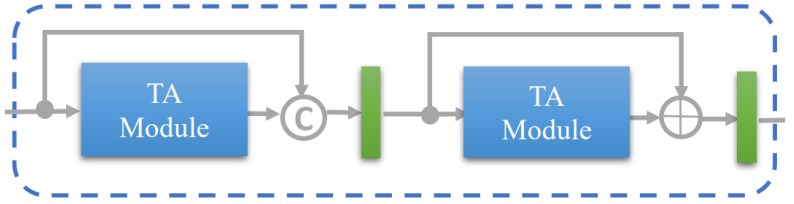}
    \caption{Structure of the stacked triple attention module.}
    \label{fig:stacked_triple_attention}
\end{figure}

To further exploit the multi-level feature attention, two triple attention modules are stacked with a structure similar to the skip connections in ResNet \cite{he2016deep} (see Fig. \ref{fig:stacked_triple_attention}). The first module takes the raw point cloud as input, while the second one works on the extracted high dimensional features. For each TA module the input is concatenated or summed to the output to fuse more feature information. Each TA module is followed by a fully connected layer to increase the feature dimension. Inside the TA modules, the attention mechanism only re-weights the features, but does not increase their dimensions.

\subsection{Pillar Feature Net}
We choose \textit{PointPillars} \cite{lang2019pointpillars} as our baseline and improve its 3D detection performance at the expense of inference time. The inference speed of \textit{PointPillars} is 42\,Hz without the acceleration of \textit{TensorRT}. Since there is a trade-off between speed and accuracy, we can further boost the accuracy by incorporating additional modules without sacrificing the inference speed too much. The pillar feature net (PFN) shown in Fig. \ref{fig:architecture}, introduced by Lang et. al, takes pillars as input, extracts pillar features, and scatters pillars back to a pseudo image for 2D convolution operations in the middle layers. The pillar feature net acts as an additional feature extractor to the stacked triple attention module. The point-wise pillar-organized features from the stacked TA module with shape ($P \times N \times C$) are fed to a set of PFN layers. Each PFN layer is simplified PointNet \cite{qi2017pointnet}, which consists of a linear layer, Batch-Norm \cite{ioffe2015batch}, ReLU \cite{nair2010rectified}, and max pooling. The max-pooled features are concatenated back to the ReLU's output to keep the point-wise feature dimension inside each pillar, until the last FPN layer. The last FPN layer makes the final max pooling and outputs a ($P \times C$) feature as the pillar feature. Pillar features are then scattered back to the original pillar location, forming a ($C \times H \times W$) pseudo image, where $H$ and $W$ are the height and width of the pillar grid. Here the location of empty pillars is padded with zeros.

\subsection{Attentive Hierarchical Middle Layers}
\begin{figure}[htbp]
    \centering
    \adjustbox{trim=0 3cm 0 0}{
        \includegraphics[width=\linewidth]{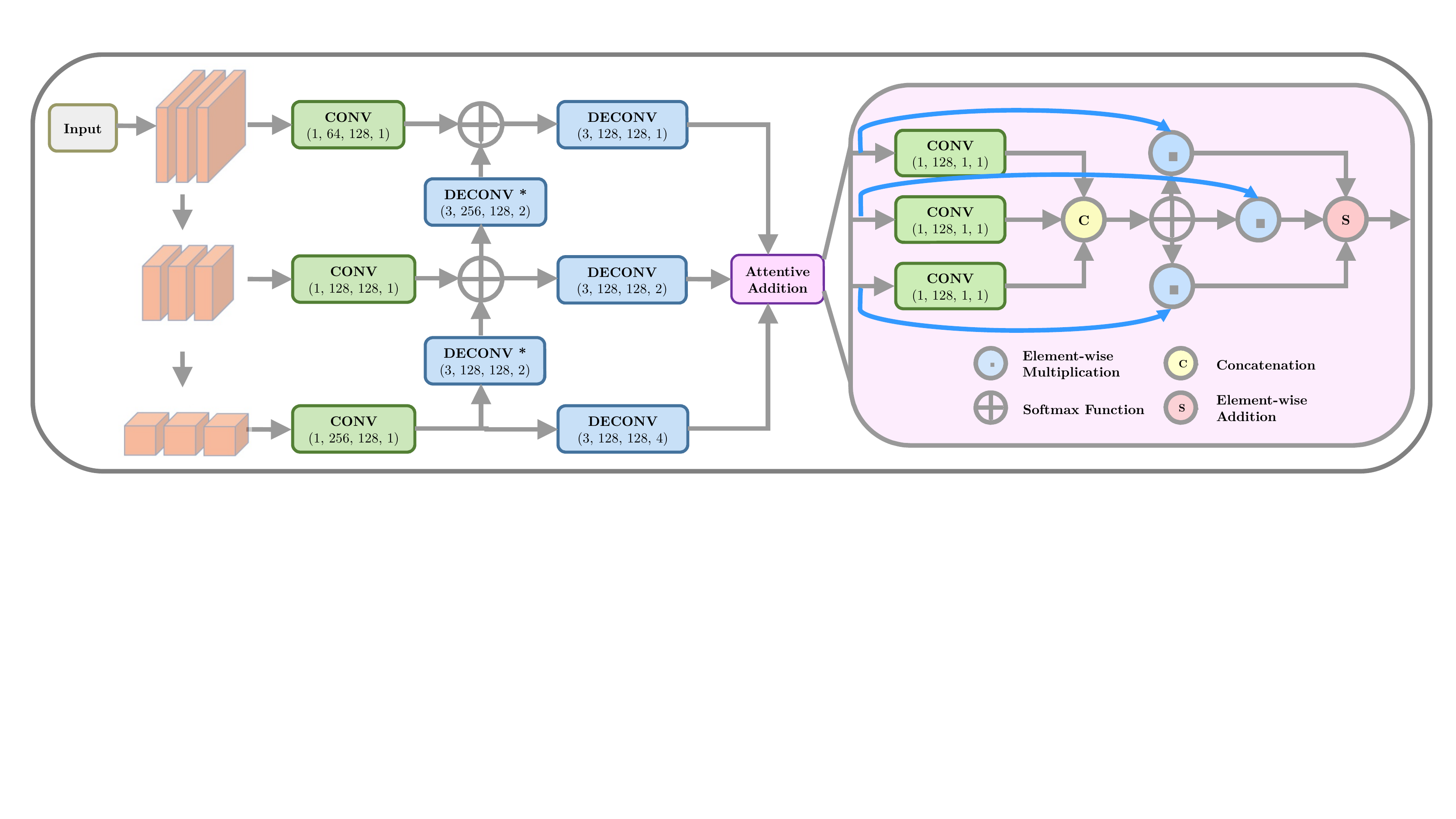}
    }
    \caption{Left: Structure of the attentive hierarchical middle layers. Right: Structure of the attentive addition operation.}
    \label{fig:attentive_hierarchical_middle_layers}
\end{figure}

We exchange the default backbone of \textit{PointPillars} with an \textit{Attentive Hierarchical Backbone} to perform 2D convolution on the pseudo image from the pillar feature net. Figure \ref{fig:attentive_hierarchical_middle_layers} depicts the structure of the attentive hierarchical middle layers. In the first stage, the spatial resolution of the pseudo image is gradually downsampled by three groups of convolutions. Each group contains three convolutional layers, where the first one has a stride of two for downsampling, and the two subsequent layers act only for feature extraction. After downsampling, deconvolution operations are applied to recover the spatial resolution. Deconvolutional layers (marked with an asterix) recover the size of feature maps with stride 2 and element-wise add them to upper branches. The remaining three deconvolutional layers make all three branches have the same size (half of the original feature map). Then the final three feature maps are combined by an attentive addition to fuse both, spatial and semantic features. The attentive addition uses the plain attention mechanism. All three feature maps are being passed through a convolutional operation and are channel-wise concatenated as attention scores. The softmax function generates the attention distribution and feature maps are multiplied with the corresponding distribution weight. The element-wise addition in the end gives the final attention output, a ($C \times H/2 \times W/2$) feature map.

\subsection{Multi-task Head}
The multi-task head outputs the final class (based on a confidence score), the 3D box position ($x, y, z$), dimensions ($l, w, h$), rotation ($\theta$) and the direction of the detected object. The direction (front/back) is being classified to solve the problem that the sine-error loss \cite{yan2018second} cannot distinguish flipped boxes. Four convolutional layers operate on the feature map separately. Figure \ref{fig:architecture} shows the brief structure of the multi-task head in the bottom right corner. One of the four heads is the IoU prediction head that predicts an IoU between the ground truth bounding box and the predicted box. It was introduced in CIA-SSD \cite{zheng2020cia} to deal with the misalignment between the predicted bounding boxes and corresponding classification confidence maps. The misalignment is mainly because these two predictions are from different convolutional layers. Based on this IoU prediction, we use the confidence function (CF) to correct the confidence map and use the distance-variant IoU-weighted NMS (DI-NMS) module post-process the predicted bounding boxes. The distance-variant IoU-weighted NMS is designed to deal with long-distance predictions, to better align far bounding boxes with ground truths, and to reduce false-positive predictions. If the predicted box is close to the origin of perspective, we give higher weights to those box predictions with high IoU. If the predicted box is far, we give relatively uniform weights, to get a more smooth final box.

\subsection{Shape-Aware Data Augmentation}
\begin{figure}[t]
    \centering
    \includegraphics[width=0.5\linewidth]{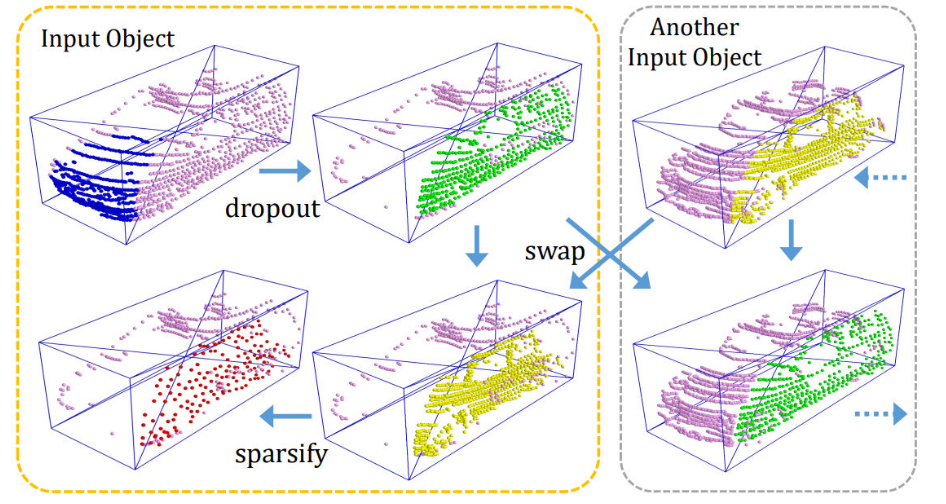}
    \caption{Illustration of the shape-aware data augmentation module \cite{zheng2021se}. It includes random dropout in
the blue subset, random swap between green and yellow subsets, and random sparsifying from the yellow to the red
subset.}
    \label{fig:shape_aware_data_augmentation}
\end{figure}

Data augmentation has proven to be an efficient way to better exploit the training dataset and help the model to be more generalized. We use the shape-aware data augmentation method proposed by SE-SSD \cite{zheng2021se} (see Fig. \ref{fig:shape_aware_data_augmentation}. This module simplifies the handling of partial occlusions, sparsity and different shapes of objects in the same class. The ground truth box is divided into six pyramidal subsets. Then we independently augment each subset using three operations, random dropout, random swap and random sparsifying.

Some traditional augmentation methods are also applied before the shape-aware augmentation, e.g. rotation, flipping, and scaling.

\subsection{Self-Ensembling Training Framework}
\begin{figure}[htbp]
    \centering
      \includegraphics[width=\linewidth]{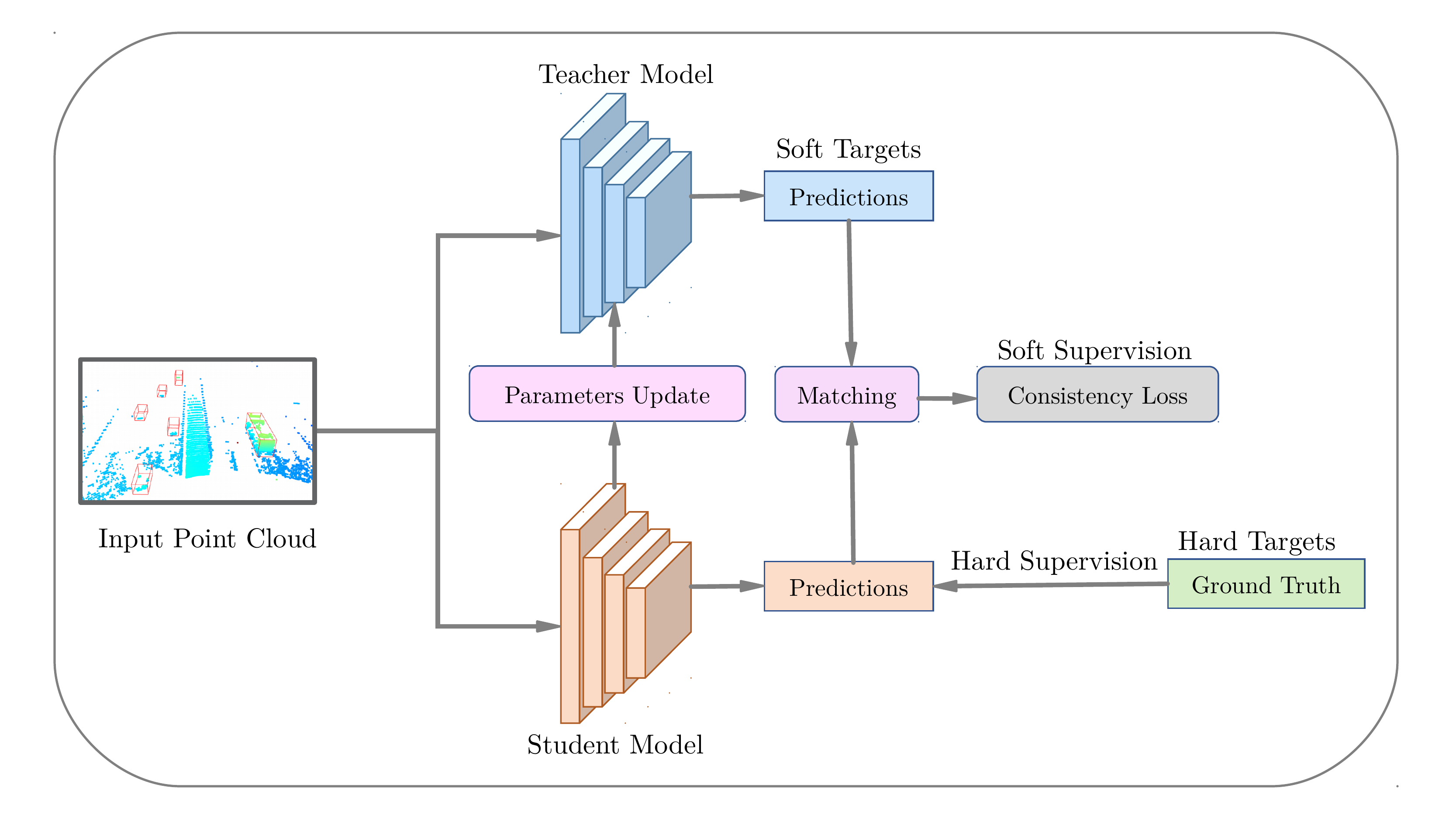}
    \caption{Self-ensembling training architecture.}
    \label{fig:teacher_student_architecture}
\end{figure}

In addition, we introduce the self-ensembling training framework \cite{zheng2021se} to do a post training: We first train the model shown in Fig. \ref{fig:teacher_student_architecture} but without the self-ensembling module, and then we take the pre-trained model as a teacher model to train the student model that has the same network structure.

The self-ensembling training architecture is show in Fig. \ref{fig:teacher_student_architecture}. The teacher model is trained with the following parameters: An adapted learning rate for each epoch was implemented with a maximum of 0.003 and a OneCycle learning rate scheduler. The model is trained for 80 epochs with a batch size of 2 for pre-training and for 60 epochs with a batch size of 4 for post-training to accelerate the training process. Adam is used as optimizer.
The predictions of the teacher model can be used as soft targets. Combined with the hard targets from the ground truth, we can provide more information to the student model. The student model and the teacher model are initialized with the same pre-trained parameters. During training, we firstly feed the raw point cloud to the teacher model and get teacher predictions. Then we apply global transformations to the teacher predictions as soft targets. For the hard targets (ground truths), we apply the same global transformations and additionally the shape-aware data augmentation. After that we feed the augmented point cloud to the student model and get student predictions. The orientation-aware distance-IoU (OD-IoU) loss is introduced in the hard supervision, to better align the box centers and orientations between the student predictions and hard targets. Comparing to the plain IoU loss \cite{iouloss}, OD-IoU loss also considers the distance and the orientation difference between two boxes. Finally, we use an IoU-based matching to match student and teacher predictions. We use the consistency loss over classification scores and bounding box predictions to provide soft supervision to the student model. The overall loss for training the student model consists of:

\begin{equation}
    \mathcal{L}_{student} =\mathcal{L}_{cls}^s + \omega_1\mathcal{L}_{OD-IoU}^s +\omega_2\mathcal{L}_{dir}^s + \lambda\mathcal{L}_{iou} + \mu_t\mathcal{L}_{consist},
    \label{eq:se_training_loss_}
\end{equation}

where $\mathcal{L}_{cls}^s$ is the focal loss \cite{lin2017focal} for box classification, $\mathcal{L}_{OD-IoU}^s$ is the OD-IoU loss for bounding box regression, $\mathcal{L}_{dir}^s $ is the cross-entropy loss for direction classification, $\mathcal{L}_{iou}$ is the smooth-$L_1$ loss for IoU prediction in the detection head, $\mathcal{L}_{consist}$ is the consistency loss, $\omega_1$, $\omega_2$, $\lambda$ and $\mu_t$ are weights of losses.

During the post-training, the parameters of the teacher model are updated based on the parameters of the student model using the exponential moving average (EMA) strategy with a weight decay of 0.999.

\section{Evaluation}
To prove the effectiveness of our proposed modules in \textit{SE-ProPillars}, we perform some ablation studies in that we stepwise include some modules into our training pipeline. We train and evaluate the model on the \textit{KITTI} dataset which is one of the most popular datasets in the autonomous driving domain. Furthermore, we train and evaluate the model on two infrastructure datasets with roadside LiDAR sensors: the \textit{IPS300+} dataset and the recently released \textit{A9} dataset. The evaluation was performed on a Nvidia GeForce RTX 3090 GPU.

\subsection{Ablation Studies}
By adding the \textit{Stacked Triple Attention} module (\textit{S-TA} module) the $mAP_{3D}$ can be increased by 1.3\% $mAP_{3D}$. This module adds 2\,ms of inference time to the baseline model and runs with 35\,ms per frame (28.57 FPS). Considering only the nearby area of the LiDAR point cloud (0-30 m), the $mAP_{3D}$ can even be increased by 8.21\% compared to 1.3\% when taking the whole area of 70\,m into account. The second module (\textit{Attentive Hierarchical Backbone}) increases the $mAP_{3D}$ again by 2.26\% compared to the baseline model.

\begin{table}[htbp]
\centering
\begin{tabular}{cccccccc}
\hline
Method &
  \multicolumn{3}{c}{3D mAP} &
  \multicolumn{3}{c}{BEV mAP} &
  Time(ms) \\ \cline{2-7}
 &
  \multicolumn{1}{c}{Easy} &
  \multicolumn{1}{c}{Moderate} &
  Hard &
  \multicolumn{1}{c}{Easy} &
  \multicolumn{1}{c}{Moderate} &
  Hard &
   \\ \hline \hline
\rowcolor{Gray}
ProPillars &
  \multicolumn{1}{c}{\underline{92.46}} &
  \multicolumn{1}{c}{80.53} &
  75.50 &
  \multicolumn{1}{c}{94.25} &
  \multicolumn{1}{c}{89.80} &
  84.98 &
  32.0 \\ \hline
naive without CF &
  \multicolumn{1}{c}{89.64} &
  \multicolumn{1}{c}{79.59} &
  74.43 &
  \multicolumn{1}{c}{93.88} &
  \multicolumn{1}{c}{89.29} &
  86.29 &
  \textbf{19.4} \\ \hline
\rowcolor{Gray}
naive &
  \multicolumn{1}{c}{89.71} &
  \multicolumn{1}{c}{79.95} &
  75.03 &
  \multicolumn{1}{c}{93.85} &
  \multicolumn{1}{c}{89.55} &
  86.75 &
  \underline{20.1} \\ \hline
naive+DI-NMS &
  \multicolumn{1}{c}{90.25} &
  \multicolumn{1}{c}{80.48} &
  75.60 &
  \multicolumn{1}{c}{94.18} &
  \multicolumn{1}{c}{89.84} &
  85.11 &
  21.4 \\ \hline
\rowcolor{Gray}
naive+DI-NMS+SA-DA &
  \multicolumn{1}{c}{92.38} &
  \multicolumn{1}{c}{\underline{80.93}} &
  \underline{76.05} &
  \multicolumn{1}{c}{\underline{96.30}} &
  \multicolumn{1}{c}{\underline{89.85}} &
  \underline{87.21} &
  22.0 \\ \hline \hline
SE-ProPillars (ours)&
  \multicolumn{1}{c}{\textbf{93.10}} &
  \multicolumn{1}{c}{\textbf{81.26}} &
  \textbf{78.17} &
  \multicolumn{1}{c}{\textbf{96.44}} &
  \multicolumn{1}{c}{\textbf{89.88}} &
  \textbf{87.21} &
  22.0 \\ \hline
  \textit{\cellcolor{cyan!10}Improvement}&
  \multicolumn{1}{c}{\textit{\cellcolor{cyan!10}+0.64}} &
  \multicolumn{1}{c}{\textit{\cellcolor{cyan!10}+0.33}} &
  \textit{\cellcolor{cyan!10}+2.12} &
  \multicolumn{1}{c}{\textit{\cellcolor{cyan!10}+0.14}} &
  \multicolumn{1}{c}{\textit{\cellcolor{cyan!10}+0.03}} &
  \textit{\cellcolor{cyan!10}+0.00} &
  \textit{\cellcolor{cyan!10}+0.00} \\ \hline
\end{tabular}
\vspace{5mm}
\caption{Ablation studies of the \textit{SE-ProPillars} 3D object detector after the pre and post training (last two rows). We report the 3D and BEV mAP of the Car and Van category (similar type enabled) on the KITTI validation set under 0.7 IoU threshold with 40 recall positions. Post-training especially improves the accuracy on the hard examples (+2.12 3D mAP).}
\label{tab:06_single_class_pretraining}
\end{table}

The result of single-class pre-training is shown in Table \ref{tab:06_single_class_pretraining} (second last row). After applying the confidence function (CF), we call this model \textit{naive SE-ProPillars} (no training tricks). We also apply the distance-variant IoU-weighted NMS (DI-NMS) and the shape-aware data augmentation (SA-DA) module in the pre-training. After enabling the multi-task head but without applying the CF, although the precision drops, the inference speed gets a large increase. The runtime is reduced from 32\,ms to 19.4\,ms. The CF in the lightweight multi-task head is trying to solve the misalignment problem between the predicted bounding boxes and corresponding classification confidence maps. After applying the confidence function, an increase is visible in all metrics, except the BEV easy difficulty keeps at the same level. After adding the DI-NMS module, the precision becomes competitive with \textit{ProPillars} considering the balance of precision and speed, although there is a relatively large gap in the 3D easy mode. The shape-aware data augmentation leads to a further increase in precision. After the pre-training our model outperforms the baseline in almost all metrics, except a tiny $0.1\%$ gap in the 3D easy difficulty. The speed of our model is much faster compared to the baseline. There is a small latency in the runtime after enabling the shape-aware data augmentation module, because as the precision increases, the computation of the DI-NMS increases.

\subsection{KITTI Dataset}
We evaluate our \textit{SE-ProPillars} detector on the \textit{KITTI} validation set using the conventional metric of KITTI, i.e. the BEV and 3D mAP result under 0.7 IoU threshold with 40 recall positions. All three levels of difficulty are taken into account. We follow the NMS parameters in \cite{zheng2021se}, set the NMS IoU threshold to 0.3, and the confidence score threshold to 0.3. We evaluate the performance on both single-class detection — Car category, and multi-class detection — Car, Pedestrian, and Cyclist. In the single-class detection, we enable the similar type, which means we train on both Car and Van categories, but treat them all as Cars. \textit{ProPillars} \cite{propillars} is used as our baseline.

\begin{figure}
    \centering
    \includegraphics[width=1.0\linewidth]{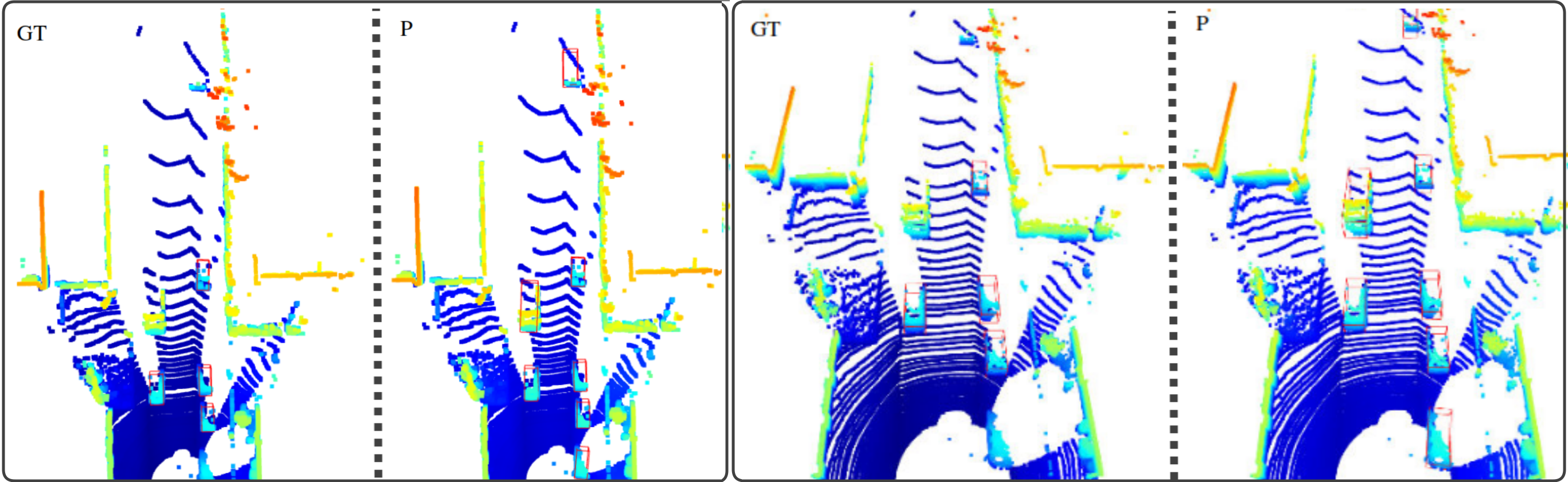}
    \caption{Left: BEV visualization of single-class detection results on the KITTI validation set. Ground truth bounding boxes are marked with GT, predicted bounding boxes with P. In KITTI, some ground truth boxes are filtered out because they are smaller than 25 pixels in height in the corresponding captured image, or completely out of the image. Some of these filtered objects can still be detected by our \textit{SE-ProPillars} which shows the good performance for detecting small objects. Right: 3D front view visualization of single-class detection results on the KITTI validation set. In this example a larger vehicle is detected as a Car.}
    \label{fig:kitti_bev_and_3d_results}
\end{figure}

We use the pre-trained model as the initial teacher model in the self-ensembling training architecture. Results are shown in Table \ref{tbl:comparison}. In the post-training, the consistency loss is enabled to provide supervision for the student model, as well as the replacement from the traditional smooth-$L_1$ loss to the orientation-aware distance-IoU loss for the bounding box regression. The SA-DA module is compulsory for providing more noisy samples to the student model. After 60 epochs of post-training, all metrics gain further improvement. Our model outperforms the baseline by $(0.64\%, 0.73\%, 2.67\%)$ 3D mAP and $(2.19\%, 0.08\%, 2.23\%)$ BEV mAP on easy, moderate, and hard difficulties respectively. At the same time, we have a much lower inference time of 22\,ms, compared to our baseline of 32\,ms. The inference time includes 5.7\,ms for data pre-processing, 16.3\,ms for network forwarding and post-processing (NMS). The hard disk used to load the model is relatively slow, which leads to a higher data loading time. The self-ensembling architecture is a training technique and therefore has no additional cost in the inference time. Some qualitative results are shown in Fig. \ref{fig:kitti_bev_and_3d_results}. We also compare our \textit{SE-ProPillars} model with other state-of-the-art detectors on the KITTI validation set (see Table \ref{tbl:comparison}).

\begin{table}[htbp]
\centering
\begin{tabular}{lcccc}
\hline
Method~~~~~~~~~~~ & ~~~~~~Year~~~~~~                                 &\multicolumn{3}{c}{3D mAP (Car)}             \\ \cline{3-5}
                                        &     &\multicolumn{1}{c}{~~~~~~~Easy~~~~~~~}            & \multicolumn{1}{c}{~~~~~~~Medium~~~~~~~}         & ~~~~~~~Hard~~~~~~~  \\ \hline \hline
\rowcolor{Gray}
VoxelNet \cite{pointrcnn}& 2018 & \multicolumn{1}{c}{82.00}          & \multicolumn{1}{c}{65.50}          & 62.90 \\ \hline
PointPillars \cite{lang2019pointpillars}&2019 & \multicolumn{1}{c}{87.75}          & \multicolumn{1}{c}{78.39}          & 75.18 \\ \hline
\rowcolor{Gray}
F-PointPillars \cite{paigwar2021frustum}&2021 & \multicolumn{1}{c}{\underline{88.90}}          & \multicolumn{1}{c}{\underline{79.28}}          & \underline{78.07} \\ \hline \hline
SE-ProPillars (ours)                    &2022 & \multicolumn{1}{c}{\textbf{93.10}} & \multicolumn{1}{c}{\textbf{81.26}} & \textbf{78.17}  \\ \hline
\textit{\cellcolor{cyan!10}Improvement}                    &\cellcolor{cyan!10}     & \multicolumn{1}{c}{\textit{\cellcolor{cyan!10}+4.20}} & \multicolumn{1}{c}{\textit{\cellcolor{cyan!10}+1.98}} & \textit{\cellcolor{cyan!10}+0.10}  \\ \hline
\end{tabular}
\vspace{5mm}
\caption{Comparison on the KITTI validation set. We show 3D multi-class detection results of SE-ProPillars and report the 3D mAP of the Car category under 0.7 IoU threshold with 40 recall positions.}
\label{tbl:comparison}
\end{table}

\subsection{IPS300+ Roadside Dataset}
Table \ref{tbl:evaluation_ips} shows the evaluation results on the full unsplit (original) IPS300+ roadside test set as well as two modified test sets: A single-LiDAR on a split IPS300+ test set and two registered LiDAR sensors on the split IPS300+ test set. The IPS300+ roadside dataset contains full-surround 3D labels whereas the KITTI dataset covers only the frontal view of the vehicle. That is why we split the IPS300+ dataset into four sub areas that represent the four different driving directions at the intersection where the dataset was recorded.

\begin{table}[htbp]
\centering
\resizebox{\textwidth}{!}{%
\begin{tabular}{lcccc}
\hline
                           & 3D mAP@0.7~~     & 3D mAP@0.5~~     & BEV mAP@0.7~~     & BEV mAP@0.5        \\ \hline \hline
\rowcolor{Gray}
1 LiDAR, split-IPS300+ & \underline{54.91}          & \underline{83.76}          & \underline{74.57}           & \underline{85.97} \\ \hline
2 LiDARs, split-IPS300+   & 63.58          & 87.52          & 78.34           & 89.42 \\ \hline
\rowcolor{Gray}
Full (unsplit) IPS300+            & \textbf{68.90}          & \textbf{97.82}          & \textbf{95.53}           & \textbf{98.16} \\ \hline
\rowcolor{cyan!10}
\textit{Improvement}   & \textit{+13.99}          & \textit{+14.06}          & \textit{+20.96}           & \textit{+12.19} \\ \hline
\end{tabular}
}
\vspace{5mm}
\caption{3D object detection results of the naive SE-ProPillars model with SA-DA. We report the 3D and BEV mAP of the Car class on a single LiDAR and two registered LiDARs on the split IPS300+ test set under 0.7 and 0.5 IoU thresholds with 40 recall positions. Using the full unsplit (original) IPS300+ dataset, we noticed a performance improvement of +13.99 3D mAP at an IoU threshold of 0.7. This shows the effect of splitting the IPS300+ point clouds into four sub-areas. Objects at the split boundary get truncated and become difficult to detect. Hence the 3D mAP values on the split datasets are lower.}
\label{tbl:evaluation_ips}
\end{table}

\begin{figure}[htbp]
    \centering
    \includegraphics[width=1.0\linewidth]{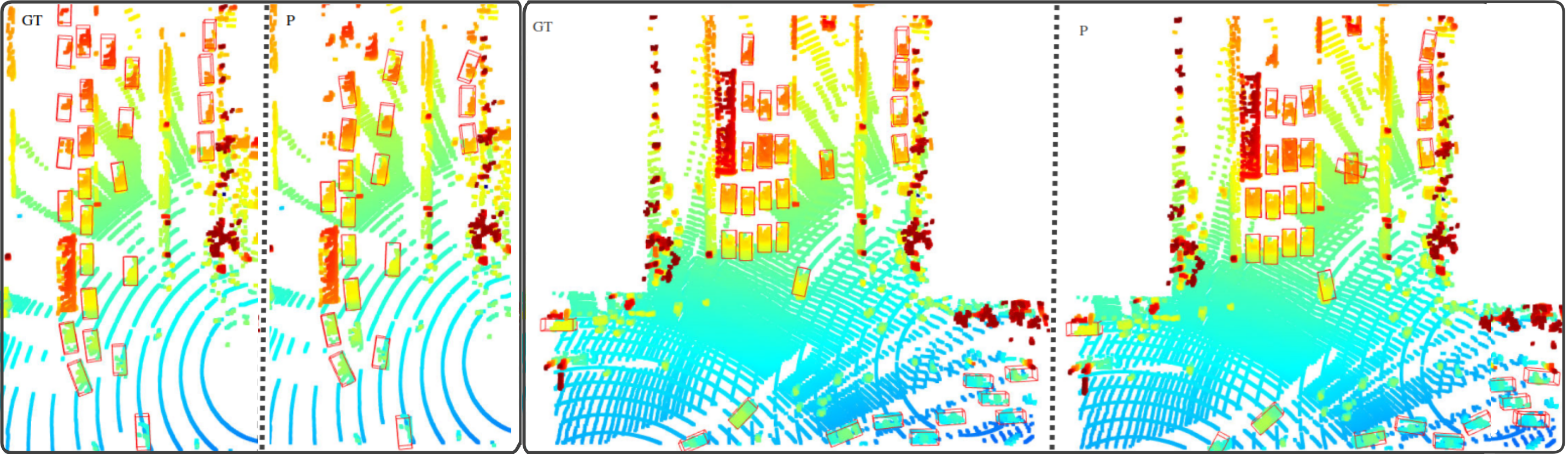}
    \caption{Left: Visualization of detection results of a single LiDAR on the split IPS300+ validation set. Ground truth bounding boxes are marked with GT and predicted bounding boxes with P. Here only 18 out of 26 vehicles were detected because of the sparse point cloud from a single LiDAR. Right: Visualization of detection results of two registered IPS300+ LiDAR point cloud scans of the IPS300+ validation set. Here 31 out of 32 vehicles could be detected and only one vehicle was missed because of the far range.}
    \label{fig:inference_results_ips_single}
\end{figure}

\subsection{A9-Dataset}
We use the \textit{naive SE-ProPillars} with SA-DA module trained on a single-LiDAR split-IPS300+ as our model, fine-tune on the training set of the A9 dataset for 100 epochs, and test on the testing set. We report our result using 3D and BEV mAP under an IoU threshold of 0.5 and 0.25 like in \cite{propillars}. The confidence score is set to a threshold of 0.1, and we again decrease the NMS IoU threshold to 0.2. This is exactly the same setting as the convention in the nuScenes \cite{caesar2020nuscenes} dataset defines. Since the LiDAR frames of the first batch of the A9 dataset are captured on the Highway A9, the distance between cars is further and there are no parked cars. We also test the case of using 0.1 as the NMS threshold. The result is shown in Table \ref{tab:06_profusion_transfer_learning}.
\begin{table}[htbp]
\centering
\resizebox{\textwidth}{!}{
\begin{tabular}{lcccc}
\hline
Metric     & 3D mAP@0.5~  & 3D mAP@0.25~       & BEV mAP@0.5~ & BEV mAP@0.25         \\ \hline \hline
\rowcolor{Gray}
ProPillars & N/A   & N/A   & N/A   & \underline{47.56} \\ \hline
Ours (NMS IoU=0.2)      & 31.03 & 48.88 & 39.91 & 49.68 \\ \hline
\rowcolor{Gray}
Ours (NMS IoU=0.1)      & 30.13 & 50.09 & 40.21 & \textbf{51.53} \\ \hline
\rowcolor{cyan!10}
\textit{Improvement} & - & - & - & \textit{+3.97} \\ \hline
\end{tabular}
}
\vspace{5mm}
\caption{Results of naive SE-ProPillars with SA-DA. We report the 3D and BEV mAP of Car on the A9 test set under 0.5 and 0.25 IoU threshold, with 40 recall positions.}
\label{tab:06_profusion_transfer_learning}
\end{table}

\section{Conclusion}
In this work we present our \textit{SE-ProPillars} 3D object detector that is an improved version of the \textit{PointPillars} model. We replace the detection head from \cite{sassd} with a multi-task head for the final prediction, that includes an IoU prediction to alleviate the misalignment between the localization accuracy and classification confidence. Attentive hierarchical middle layers were proposed to perform 2D convolution operations on the pseudo image. The attentive addition operation concatenates the hierarchical feature maps. We add two training techniques to our baseline: the shape-aware data augmentation module and the self-ensembling training architecture to improve the accuracy without adding additional costs in the inference time. We show the generalization ability of our model and make it more robust to noise by applying multiple data augmentation methods. Sufficient data collection is key to achieve a good accuracy. That is why we generated a synthetic dataset proSynth with 2k labeled LiDAR point cloud frames and trained our model on that. We do several sets of experiments for each module to prove its accuracy and runtime performance in different ablation studies. Finally, we evaluate our 3D detector on different datasets that shows that our model significantly ourperforms the baseline model.

\section{Future Outlook}
Our \textit{SE-ProPillars} 3D object detector is a significant contribution within the area of roadside LiDAR-based 3D perception and can be used for smart city applications in the future. A far-reaching view can be provided to connected and automated vehicles and detected vehicles around the corner can be communicated to all traffic participants to improve road traffic safety. LiDAR sensors can be used for anonymous people and vehicle counting to calculate the traffic density in real-time. Traffic monitoring will improve traffic flow and increase public safety. Anonymous 3D detections will be sent to connected vehicles to improve path and maneuver planning. Detecting edge cases and rare events like accidents or breakdowns is part of the future work to notify emergency vehicles immediately. The used LiDAR sensors are part of an Cooperative Intelligent Transport System (C-ITS), that can share dangers on the road (e.g. vehicle breakdowns or falling objects) with all connected vehicles in real-time. Using these LiDAR sensors in combination with smart traffic lights will improve the traffic flow in the future by enabling a green wave for traffic participants (e.g. cyclists). Connecting roadside LiDARs with vehicles will lead to a safe and efficient driving and accidents will be prevented before they happen. Connected and cooperative vehicles will be able to make safer and more coordinated decisions and improve their path and maneuver planning in real-time. Finally, an online and anonymized database of traffic participants can be created and shared with connected traffic participants to automatically react to potential incidents. This platform will share all traffic data and stream live sensor data to improve public safety.

\section*{Acknowledgements}
This work was funded by the Federal Ministry of Transport and Digital Infrastructure, Germany as part of the research project Providentia++ (Grant Number: 01MM19008A). The authors would like to express their gratitude to the funding agency and to the numerous students at TUM who have contributed to the creation of the first batch of the A9-Dataset.

\bibliographystyle{spmpsci.bst}

%
%

\end{document}